\documentclass[a4paper,twoside]{article}
\usepackage{epsfig}
\usepackage{booktabs}
\usepackage{subcaption}
\usepackage{calc}
\usepackage{amssymb}
\usepackage{amstext}
\usepackage{amsmath}
\usepackage{url}
\usepackage{graphicx}
\usepackage{amsthm}
\usepackage{multicol}

\usepackage{pslatex}
\usepackage{apalike}
\usepackage[bottom]{footmisc}

\usepackage{SCITEPRESS}     % Please add other packages that you may need BEFORE the SCITEPRESS.sty package.
\usepackage{amsmath,amssymb,amstext}

\usepackage{graphicx} % For including graphics 
\usepackage{float}
\usepackage{nomentbl} 
\makenomenclature 

\usepackage{ifpdf}
\usepackage{amsthm}
\usepackage{longtable}

\usepackage{algorithm}
\usepackage{algpseudocode}

\usepackage{caption}
\usepackage{subcaption}

\usepackage{derivative}
\usepackage[\ifpdf pdftex,\fi,pagebackref=false]{hyperref} % with basic options
\begin{document}

\title{{FlexPooling with Simple Auxiliary Classifiers in Deep Networks}
}

% \author{\authorname{First Author Name\sup{1}\orcidAuthor{0000-0000-0000-0000}, Second Author Name\sup{1}\orcidAuthor{0000-0000-0000-0000} and Third Author Name\sup{2}\orcidAuthor{0000-0000-0000-0000}}
% \affiliation{\sup{1}Institute of Problem Solving, XYZ University, My Street, MyTown, MyCountry}
% \affiliation{\sup{2}Department of Computing, Main University, MySecondTown, MyCountry}
% \email{\{f\_author, s\_author\}@ips.xyz.edu, t\_author@dc.mu.edu}
% }
\author{\authorname{Muhammad Ali\sup{1}\orcidAuthor{0000-0001-9320-2282}, Omar Alsuwaidi\sup{1}\orcidAuthor{0000-0000-0000-0000} and Salman khan\sup{1}\orcidAuthor{0000-0002-9502-1749}}
 \affiliation{\sup{1}Department of Computer Vision, Mohamed bin Zayed University of Artificial Intelligence (MBZUAI), Abu Dhabi, UAE}
\email{\{muhammad.ali, omar.alsuwaidi\}@mbzuai.ac.ae, salman.khan@mbzuai.ac.ae}
 }

\abstract{In Computer Vision, the basic pipeline of most convolutional neural networks (CNNs) consists of multiple feature extraction processing layers, wherein the input signal is downsampled into a lower resolution in each subsequent layer. This downsampling process is commonly referred to as pooling, an essential operation in CNNs. It improves the model’s robustness against variances in transformation, reduces the number of trainable parameters, increases the receptive field size, and reduces computation time. Since pooling is a lossy process yet crucial in inferring high-level information from low-level information, we must ensure that each subsequent layer perpetuates the most prominent information from previous activations to aid the network’s discriminability. The standard way to apply this process is to use dense pooling (max or average) or strided convolutional kernels. In this paper, we propose a simple yet effective adaptive pooling method, referred to as FlexPooling, which generalizes the concept of average pooling by learning a weighted average pooling over the activations jointly with the rest of the network. Moreover, attaching the CNN with Simple Auxiliary Classifiers (SAC) further demonstrates the superiority of our method as compared to the standard methods. Finally, we show that our simple approach consistently outperforms baseline networks on multiple popular datasets in image classification, giving us around a 1-3\% increase in accuracy.}

\onecolumn \maketitle \normalsize \setcounter{footnote}{0} \vfill

\section{\uppercase{Introduction}}
\label{sec:introduction}

In the current Computer Vision community, the vast majority of image label learning techniques rely on the hard classification of samples in order to learn a function mapping between the images and their corresponding classes. Usually, under a supervised training setting, class-label learning involves using cross-entropy and binary cross-entropy loss functions for multiclass and binary class classification, respectively. The choice for the hypothesis class to learn the label function mapping is most often parameterized as a convolutional neural network (CNN) with multiple processing and downsampling layers. CNNs are very suitable for such a task as they mimic our natural ability to perceive images by dividing an image into many small sub-images and processing them \emph{locally} for feature extraction, one by one. They also utilize parameter sharing via learned convolutional kernels based on the assumption that learning a meaningful pattern in one part of the image may also be helpful in another part of the image, and that is especially true in the early stages of the CNN, where the features learned by the convolutional kernels tend to be less abstract and more generalizable. Moreover, thanks to their intrinsic design, they can gradually reduce the dimension of the input, allowing for faster processing, reduction in the number of parameters, and increased receptive field size, which in turn allows the model to learn high-level information from low-level ones. All these properties have made CNNs very robust and efficient models when dealing with image processing.

Another building block of almost all convolutional neural networks (CNNs) is a crucial technique known as pooling, a process that significantly downscales the incoming image by preserving a small portion of the feature map, which resembles the most important pixels for the task solution. A usual pipeline of a CNN is composed of multiple stages, where each stage involves a series of feature extraction (convolutional) layers, followed by a pooling layer, where the image gets downscaled further in each consecutive stage.

Pooling allows the CNN model to be invariant to tiny distortions and further perturbations in the image. It also dramatically enlarges the receptive field between intermediate and output nodes while lowering the computational cost and parameter count, similar to what a convolutional kernel does but more amplified  \cite{Papineni2001BLEU}. Out of the possible pooling methods, Average or mean pooling and Max pooling are the two most frequently used pooling techniques in various CNN architectures like VGG \cite{KarenSimonyan}, ResNet \cite{HeDeepRecognition}, and GoogLeNet \cite{Szegedy2015GoingConvolutions}.

Despite their popularity and widespread use, average pooling and max pooling possess serious drawbacks and model limitations. For one, despite the entire ensemble of parameters in a CNN being trained jointly from end to end, the average and max pooling layers, by definition, are unlearnable and hence do not adapt well to seen data during training nor generalize effectively to unseen data \cite{gholamalinezhad2020pooling}. Furthermore, since pooling is a lossy process that discards some information from the incoming data, we need to ensure that our pooling process extracts the most prominent and relevant information for the end task solution. However, average and max pooling both rely on an \emph{a priori} assumptions: the mean of locally neighboring pixels is a good representation of the region, and the maximum response from a group of neighboring pixels is the most relevant representation in that region, respectively. While these assumptions have some theoretical justification and can be worked around via learnable convolutional kernels, they can easily hinder the network's ability to learn efficiently and serve as a bottleneck because of their incompatibility in adapting to activations due to lack of learnability.
More recent CNN designs, like ResNets, frequently apply convolutions with strides greater than one as a replacement for pooling layers. However, strided convolutions come with their own disadvantages and fail to act as an actual pooling process. Conversely to traditional pooling layers, they \textbf{no} longer treat and pool each feature map \emph{independently}; instead, they aggregate all the feature maps channel-wise to get the resulting output activation. Lacking the ability to pool each feature map independently goes against the inherent design of CNNs, where the generation of each feature map in a stack of feature maps is the result of a \emph{single} convolutional kernel that is \emph{independent} of other convolutional kernels which generated the rest of the feature maps in the same stack. Hence, each feature map represents a \emph{unique distribution} of locally extracted features that might be similar but independent of the rest of the feature maps in the same stack. That is why it is of utmost importance to treat and pool each feature map individually in order to be able to extract and perpetuate the most meaningful compact representation during the lossy process of pooling.

However, in recent years and with the advancements of CNN architectures, the most common use case of downsampling via pooling has been the use of Global Average Pooling, which holistically averages the feature maps individually across the height and width dimensions in the last convolutional stage before feeding the outcome into the projection head in order to make the class prediction \cite{Ren2015FasterNetworks}. Global average pooling offers multiple advantages over the traditional way of using fully connected layers. It reduces the number of trainable parameters, also enforcing correspondences between feature maps and classes. Therefore, allowing us to interpret the final processed feature maps as class confidence maps. Despite these significant benefits of using global average pooling over fully connected layers, they still suffer from the same limitations as the standard average pooling that were discussed earlier in this paper, mainly incompatibility in adapting to activations due to lack of learnability, which can hurt the model's ability to generalize to unseen data.

In this work, we primarily focus on improving this vital yet often overlooked process of feature map to class correspondence via global pooling. We aim to replace the global average pooling layer by substituting it with a more robust flexible pooling layer, named FlexPool, that is trainable jointly with the network in an end-to-end fashion and is fully differentiable. FlexPool aims to learn the best set of weights in a weighted average for \emph{each} feature map in the last stage of convolutions, allowing the model to \emph{learn} the appropriate correspondences between feature maps and categories. Having the ability to learn a weighted average global pooling over the feature maps with the rest of the network, FlexPool is an effortless yet efficient adaptive pooling technique that generalizes the idea of global average pooling. Additionally, attaching the CNN with Simple Auxiliary Classifier (SAC) heads along different convolutional stages further demonstrates our method's superiority over standard global average pooling.

We demonstrate that substituting FlexPooling as the global pooling layer of choice improves performance on the different datasets using different ResNet sizes. With the introduction of this simple but effective technique, we propose using it with any existing state-of-the-art CNN to get improved results easily. In order to further explain the concept sequentially, we organized the rest of the paper as follows; Section 2 reviewed the related work done on pooling layers. In Section 3, we present our suggested model and dictate its formulation, explaining the effects of FlexPooling and its variants. We describe the experiments and analyze the outcomes of those experiments in detail in Section 4 using different benchmark datasets and models. Finally, Section 5 concludes with a summary of our work.

\section{\uppercase{Materials and Methods}}

Among the two most popular pooling techniques (average and max pooling), max pooling tends to outperform average pooling in terms of discriminability in most CNN architectures and tasks, especially for features with low activation intensities, as studied by \cite{Boureau2010ARecognition} when examining the different pooling techniques. Max pooling seeks to preserve the most important details because it is crucial for the network's ability to discriminate \cite{FPAC}. However, because only one node is chosen in each local neighborhood, this can cause the gradient flow to experience a disruption in its gradient magnitudes in connections branching out from that node during the backward pass of the CNN. Max pooling also produces relatively sparse results when downscaling images \cite{Gulcehre2013Learned-NormNetworks}. Moreover, the fact that the appropriate choice of pooling relies upon the CNN structure and dataset distribution is another drawback of traditional pooling layers, which requires the practitioner to perform extensive empirical testing to determine the appropriate pooling technique for the specific task solution.

Since the two most commonly used methods for pooling both have their own advantages and limitations, one might suspect that the best way to extract meaningful information during pooling may fall somewhere between the two methods of average and maximum pooling.
Previous research on pooling techniques has focused on using unusual pooling ratios or altering the receptive field size of the pooled region. 

Researchers like \cite{ionescu2015training} offered a strategic approach to make it possible to include deeper networks with higher-order pooling layers. Performing inter-channel max pooling is advised by Maxout \cite{Goodfellow2013MaxoutNetworks}, while a fractional downscaling ratio is used in fractional pooling \cite{GholamalinezhadPoolingReview}, which results in a more steady size reduction. Low-pass filtering was implemented by \cite{Rippel2015SpectralNetworks} to downsample feature maps in spectral space. As details are primarily concentrated in higher frequencies, this smoothes the input rather than maintaining them.
Researchers also worked on various receptive field sizes of pooled pixels\cite{Rippel2015SpectralNetworks}. Ionescu et al. offer a strategy that makes it possible to include deeper networks with higher-order pooling layers,
 As details are primarily focused on higher frequencies, this smoothens the input rather than maintaining them \cite{SaeedanDetail-PreservingNetworks} through learnable pooling layers, recent research has sought to maintain the most discriminative parts of the data/activations and eliminate the redundant ones. There have been several attempts to combine max and average pooling\cite{GholamalinezhadPoolingReview}; nonetheless, none truly successfully harnessed the advantages in both \cite{GholamalinezhadPoolingReview}. Other pooling techniques included scaling according to the input size \cite{He2014SpatialRecognition}, enabling CNNs to handle a range of image sizes.

In study \cite{SaeedanDetail-PreservingNetworks}, researchers  proposed a method to preserve the subtle elements within an input image that are essential in representing an accurate visual impression. Its purpose was to retain the refined details typically lost by the average pooling and have it improved along the network in the learning path, thereupon helping us to get the relevant missing information. This approach motivates this work in attempting to preserve the fine pixel-specific details and avoid losing any information as much as possible; thus, we replace the existing standard average pooling with our novel FlexPooling layer
\section{Methodology}
\subsection{FlexPool}

FlexPooling, is a revolutionary trainable pooling strategy which is inspired by \cite{OVGO} and  \cite{OVLEARN}, who found that overparameterized neural networks converge faster and generalize better to various datasets. It makes average pooling more general by learning a weighted average pooling over the latent feature activations of the network at the same time. FlexPool improves feature map to class correspondence by using parameters trained jointly end-to-end
We Replace FlexPooling's global average pooling layer with one that is more robust, flexible, and trainable from beginning to end. In the last step of convolutions, FlexPool learns the best weights for a weighted average sum for each feature map. This helps the model understand how feature maps and categories are linked. This basic version is described in Figure 1.

\begin{figure*}[h]
   \centering
    \includegraphics[scale=0.25]{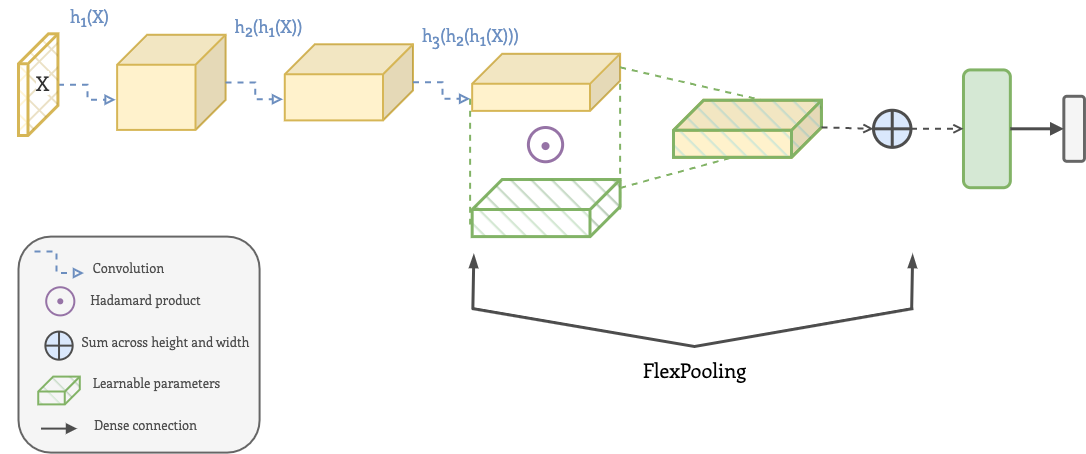} 
    \caption{\textbf{Proposed Method:} After passing the image $X$ through a series convolutional layers in ResNet20 we initialize a FlexPooling block composed of learnable parameters that matches the shape of the feature maps in the last stage. Those parameters are initialized to resemble global average pooling layer (the initialized value for each parameter is $\frac{1}{height \times width}$). Finally, we take the dot product of the two blocks and sum across the height and width dimensions, resulting in the final flattened layer before the projection head is applied, yielding the class predictions in gray.}
    \label{fig: flex}
\end{figure*}

Our methodology consists of two main parts; firstly, we introduce the concept of the trainable pooling layer, FlexPooling, which aids in the convergence of our objective function. Secondly, we augment the CNN implementation with SACs to further demonstrate the effectiveness of our FlexPooling technique compared to traditional approaches like global average pooling. SACs do not require any additional processing units like dense or convolutional layers; instead, they immediately take the feature map representation (block) at any particular stage, then collapse it into a flattened vector through a global pooling approach. The detailed description of two approaches are summarized in Figure \ref{fig: flex} and Figure \ref{fig: flexsac}, respectively.
First, we present and detail FlexPooling, a novel trainable pooling technique inspired by the works and findings of \cite{OVGO}, \cite{OVLEARN}, %and \cite{overpara},
whom all concluded that overparameterized neural networks achieve easier convergence and generalize better to different datasets. It serves as a simple yet effective adaptive pooling method that generalizes the concept of average pooling by learning a weighted average pooling over the network's latent feature activations jointly with the rest of the network. FlexPool primarily focuses on improving the vital yet often overlooked process of feature map to class correspondence via a global pooling operation. In FlexPooling, replace the global average pooling layer with a more robust and flexible pooling layer that is trainable jointly with the rest of the network in an end-to-end fashion and is fully differentiable. FlexPool aims to learn the best set of weights in a weighted average sum for \emph{each} feature map in the last stage of convolutions, allowing the model to better \emph{learn} the appropriate correspondences between feature maps and categories.
For any given input image $X$ that is passed through the convolutional operation for feature extraction, the resulting feature map is a processed and downsampled version of the input image. As we process deeper through the CNN, the feature map representation of the input image continues to shrink along the height and width dimensions (size) while enlarging along the channel dimension (depth). During the last stage of processing, a global average pooling layer is applied to each individual feature map resulting in a single output for each feature map as a flattened vector. This global pooling helps the CNN establish a correspondence between feature maps and classes as class confidence maps. Finally, the flattened vector is sent to the network's project head, which usually consists of a series of non-linearly activated dense layers, before being sent to the softmax function to generate class predictions. This setup is the most adopted setup when using CNNs. The resulting size of a feature map after being processed by a convolution operation is given by the following \cite{CS230}:
\begin{equation}
    \frac{\left(N - K + 2P\right)}{S} + 1
\end{equation}
where $N$ is the feature map's input volume, $K$ is the convolution's kernel size, $P$ is the number of padding, and $S$ represents the stride of the kernel.

In addition to the traditional approach, during the last stage of processing, we replicate the shape of the final feature maps to initialize our FlexPooling block such that it matches the final stage's block shape. The FlexPooling block consists of trainable parameters, each initialized to resemble the value of the global average pooling layer. Hence, every value in the FlexPooling block initially is set to $\frac{1}{height \times width}$. Afterward, the dot product is taken between the FlexPooling block and the final stage's processed block to obtain a weighted block. Finally, we sum across the size of the weighted block, yielding a weighted average output for each feature map. The objective is to have these values be trainable parameters in such a manner that it aids in the model learning process during training. However, nothing guarantees that the FlexPool blocks' parameters remain as weighted average values. In fact, due to the stochasticity and the noise in the training process, it very well might be the case that some of the parameters can take on negative values, thus, no longer representing a weighted average. To tackle this issue, we propose to \emph{regularize} the FlexPooling parameters $\left(\textbf{w}^{fp}\right)$ by introducing a regularization term $R\left(FlexPool\right)$ in such a way that enforces a weighted average when summing the parameter values across the size of the FlexPooling block. This parameter regularization can be accomplished in a couple of ways. In our experiments, we utilize the Mean Square Error (MSE) difference to enforce the desired behavior on the FlexPool parameters:

\begin{equation} \label{flex_reg}
    R(FlexPool) = \sum_{k=1}^C\left(\sum_{i=1}^N\sum_{j=1}^N w_{i, j, k}^{fp} - 1\right)^2
\end{equation}
where $N$ is the FlexPool's volume size, $C$ is the number of channels in the FlexPool block, and $w_{i, j, k}^{fp}$ is the weight in the $i^{th}$ row, $j^{th}$ column of the $k^{th}$ channel in the FlexPool block.

\begin{table*}[]
\centering
\caption{\textbf{FlexPooling:} Performance comparison on benchmark datasets, including ImageNet, CIFAR10, Fashion-MNIST, and CIFAR100. ResNet20 is used as the baseline of this experiment, though any model can be used. It is apparent across all datasets that adopting FlexPool yields optimal results as compared to the standard average pooling. An increased accuracy trend emerges when moving across the columns of each row in the table, demonstrating the consistency and effectiveness of FlexPooling.}
\label{tab:my-t}
\begin{tabular}{@{}cccclcl@{}}
% \multicolumn{7}{c}{\cellcolor\textbf{}}
%                                     \\ \midrule
\textbf{Dataset Methods} & \textbf{AvgPool} & \textbf{FlexPool} & \multicolumn{2}{c}{\textbf{FlexPool + Reg}} & \multicolumn{2}{c}{\textbf{FlexPool + Reg + Dropout}} \\ \midrule
\textbf{CFAR10}       & 91.63 & 91.91 & \multicolumn{2}{c}{91.87} & \multicolumn{2}{c}{\textbf{92.32}} \\ \midrule
\textbf{Fashion-MNIST} & 91.38 & 92.35 & \multicolumn{2}{c}{93.48} & \multicolumn{2}{c}{\textbf{93.98}} \\ \midrule
\textbf{CFAR100}      & 66.18 & 66.50 & \multicolumn{2}{c}{66.73} & \multicolumn{2}{c}{\textbf{68.30}} \\ \midrule
\textbf{ImageNet}     & 46.50 & 47.03 & \multicolumn{2}{c}{48.11} & \multicolumn{2}{c}{\textbf{48.55}} \\ \bottomrule
\end{tabular}
\end{table*}

\begin{figure}[h]
   \centering
    \includegraphics[scale=0.3]{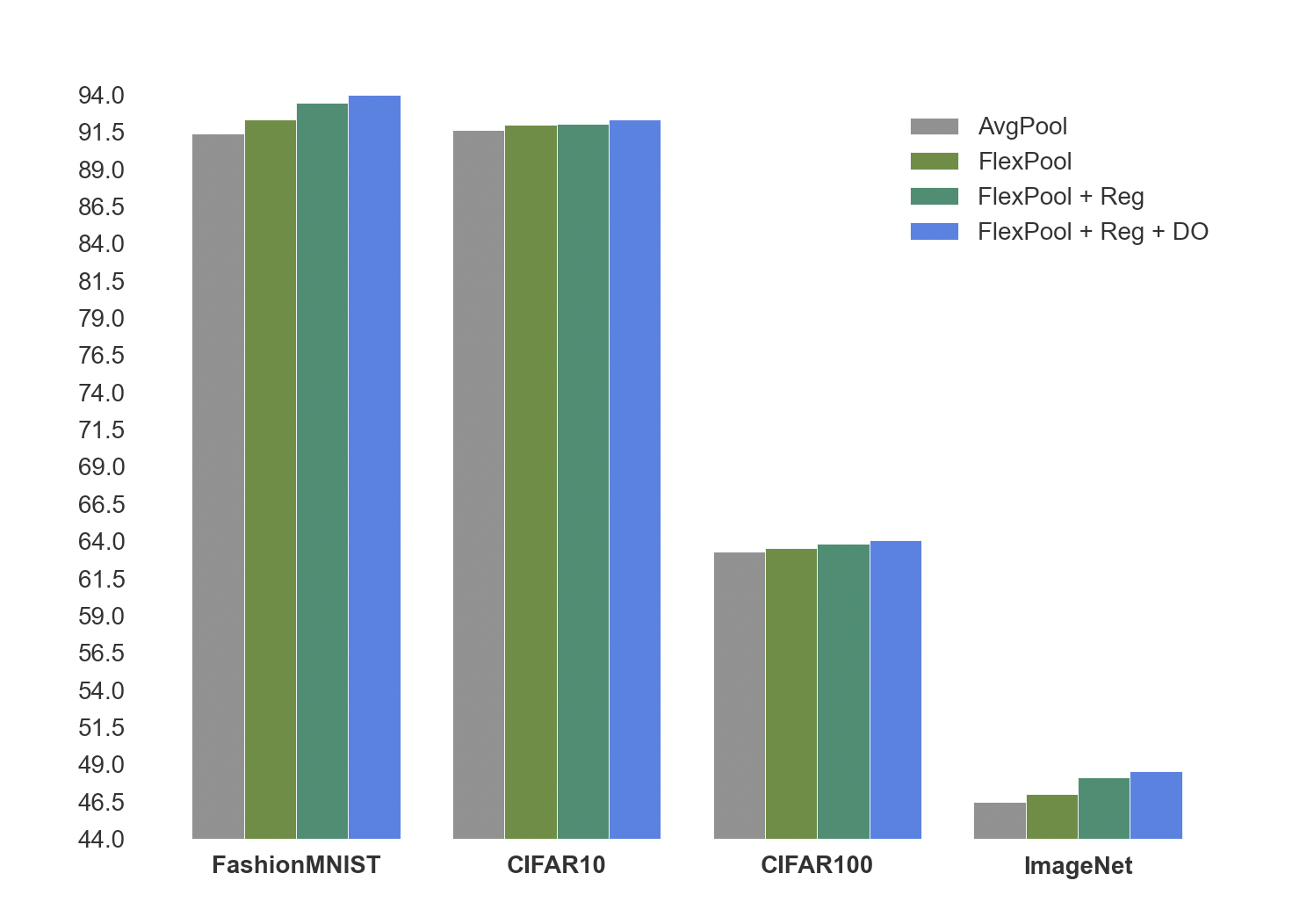} 
    \caption{FlexPooling performance on different benchmark datasets. As given in Table 1, these plots show consistent improvement in accuracy with the introduction of FlexPooling and its variants.}
    \label{fig: flex table}
\end{figure}
\subsection{FlexPool with SAC}

\begin{figure*}[h]
   \centering
    \includegraphics[scale=0.3]{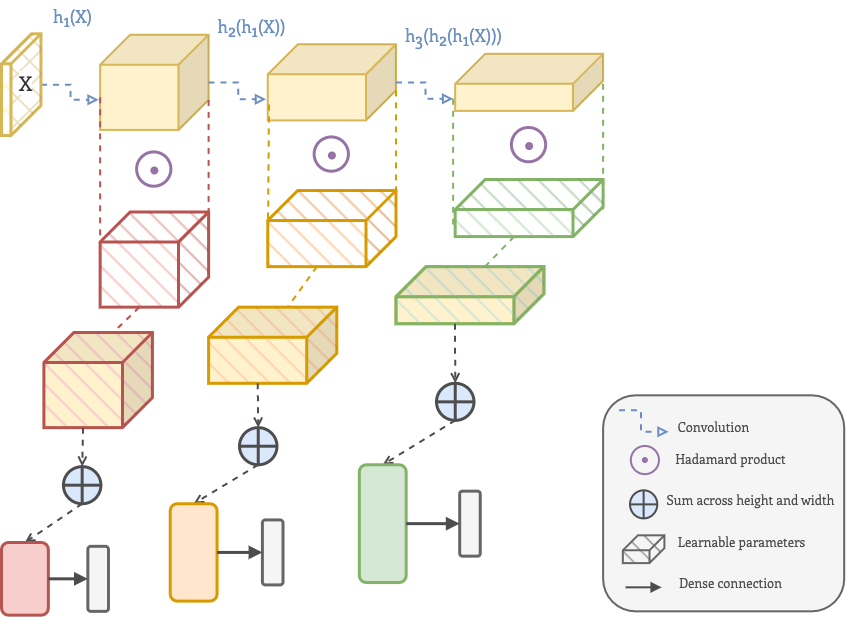} 
    \caption{\textbf{FlexPooling with SACs approach}. The cross-entropy loss is calculated for each class prediction (in gray) after every FlexPooling block and is taken as a weighted sum to obtain the final loss. The loss values obtained from different stages of the CNN aid the network in learning more abstract concepts by boosting the gradient signal throughout the entire network.}
    \label{fig: flexsac}
\end{figure*}

Here our suggested network takes feature maps from early stages and compress them into linear representation via FlexPooling, where no convolutional processing is involved,yet we downsample the feature maps into single pixel representation.

Compared to basic FlexPooling, here we adopt FlexPooling with the attachment of Simple Auxiliary Classifiers(SAC) onto the CNN to further observe the effects of pooling feature maps at different stages on accuracy. The loss values at each stage of the CNN are aggregated to obtain a weighted average loss. Particular weights $\lambda1$, $\lambda2$, and $\lambda3$, provided as hyperparameters, are assigned to each loss function that emerged from each stage of the CNN. We note that the following condition of the loss weights must hold to achieve stable learning: $\lambda_1 < \lambda_2 < \lambda_3$. The final weighted average loss is given by:

\begin{multline}
    Loss = \lambda_1 L\left(h_1\left(x\right),y\right) + \lambda_2 L\left(h_2\left( h_1\left(x\right)\right),y \right) \\+ \lambda_3 L\left(h_3 \left(h_2\left(h_1\left(x \right)\right)\right),y \right)
\end{multline}

where the loss function $L$ used is the categorical cross-entropy loss\cite{Zhang}, given by:

\begin{equation}
L(\left(h\left(x\right),y\right)) = -\sum_{c=1}^My_{o,c}\log(h_{o,c})
\end{equation}

Where $M$ is the number of classes, $h$ is predicted probability observation of class $c$, and $y$ is true label assigned to class $c$. In our experiments, the weights corresponding to the loss functions from each stage were set to $\lambda_1=0.1$, $\lambda_2=0.2$ and $\lambda_3=0.7$. After the aggregate loss is computed from weighted cross entropy losses at different stages, we back propagate the loss through the entire network to update the model parameters, including FlexPool's parameters. We evaluate our approach on various benchmark datasets in image classification, to demonstrate its effectiveness.

\begin{table*}[]
\centering
\caption{\textbf{FlexPooling wih SAC's:} Performance comparison on benchmark datasets, including ImageNet, CIFAR10, Fashion-MNIST, and CIFAR100. ResNet20 is used as the baseline of this experiment, though any model can be used. Significant improvement is exhibited in all datasets when going from AvgPool to FlexPool. While the best improvement is shown in ImageNet, a consistent improvement in accuracy is also demonstrated in the other datasets.}
\label{tab:my-table}
\begin{tabular}{@{}ccccc@{}}
% \multicolumn{7}{c}{\cellcolor\textbf{}}
                                    \\ \midrule
\textbf{Datasets/Methods} & \textbf{AvgPool} & \textbf{FlexPool} & \textbf{FlexPool + Reg} & \textbf{FlexPool + Reg + Dropout} \\ \midrule
\textbf{CIFAR10}      & 89.31\% & 91.61\% & 91.86\% & \textbf{92.03}\%  \\ \midrule
\textbf{Fashion-MNIST} & 93.55\% & 93.65\% & 93.48\% & \textbf{93.84}\% \\ \midrule
\textbf{CIFAR100}      & 63.79\%& 64.12\% & 64.53\% & \textbf{65.57}\% \\ \midrule
\textbf{ImageNet}      & 45.13\% & 47.55\% & 48.64\% & \textbf{49.01}\%\\ 
 \bottomrule
\end{tabular}
\end{table*}

\begin{figure}[h]
   \centering
    \includegraphics[scale=0.25]{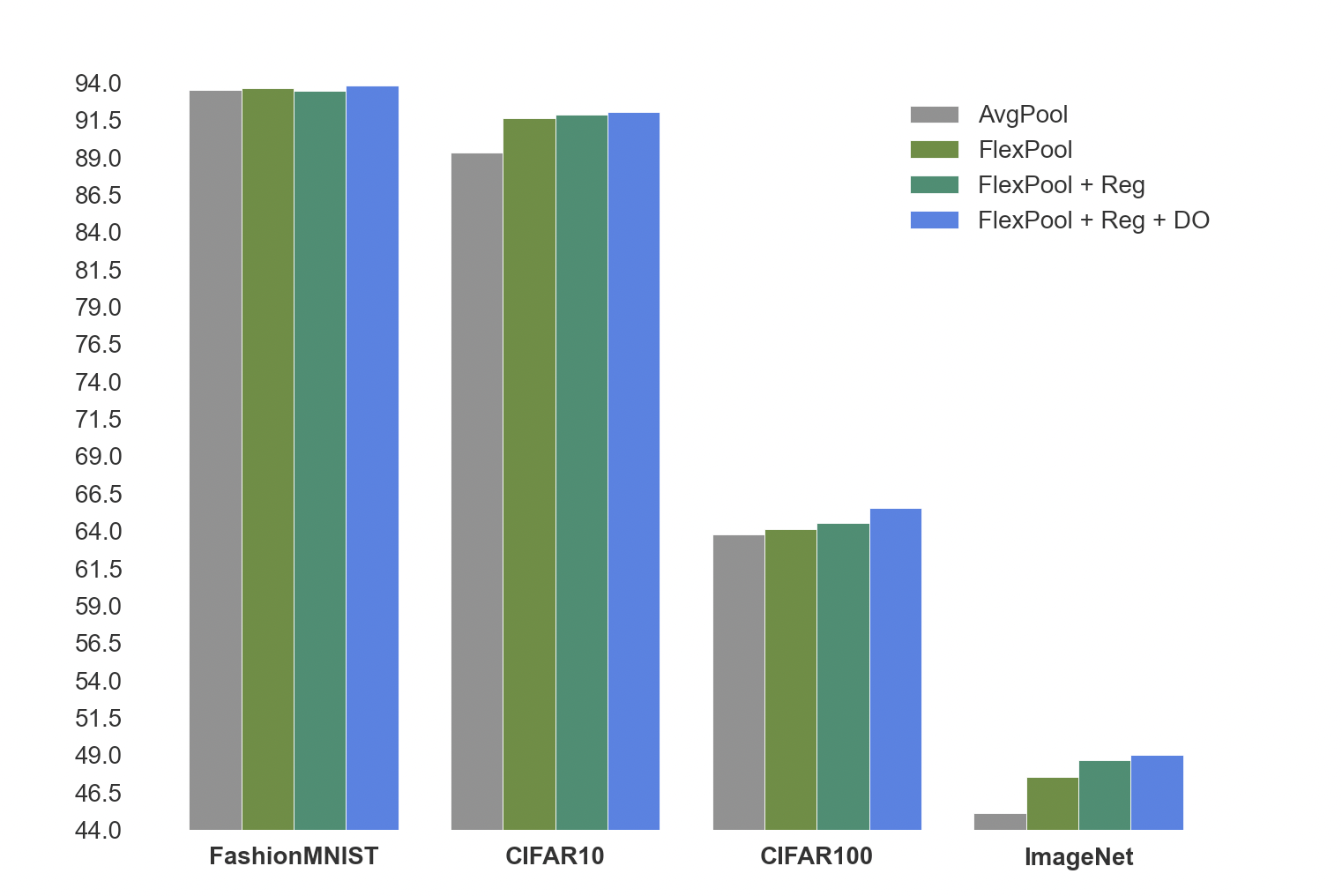} 
    \caption{FlexPooling with SAC performance on different benchmark datasets. As given in Table 2, these plots show further consistent improvement in accuracy with the introduction of FlexPooling and its variants.}
    \label{fig: SAC flex table}
\end{figure}

\section{Results and Discussions}
\label{sacflex}

We evaluate our proposed classifier with FlexPooling, on several datasets including, Tiny ImageNet\cite{IMAGNet}, CFAR100\cite{CFAR10}, and also very small scale datasets including  CFAR10\cite{CFAR10} and  FashionMNIST\cite{FMNIST}
and compare it with standard classifier using average pooling.
We conduct extensive evaluation study to evaluate the effect of different components on the performance of our model. We use accuracy as the evaluation metric for performance evaluation of our approach. We conduct experiments in two settings, including with FlexPooling only, FlexPooling with auxiliary classifiers. In each setting we evaluate different combinations: FlexPool only, FlexPool with regularization and FlexPool with regularization and droput.  

\subsection{Experimental Setup and implementation Details }

 We take the input image of size 32x32 in the case of tiny ImageNet and pass it through the convolutional blocks of ResNet20 in our case, we may use any other network. We apply our method in two settings : in the single classifier settings an image x is pass through different convolutional blocks and  we apply FlexPooling at the end before feeding it to the linear classifier and compute the relevant cross entropy loss. For the FlexPool  using SAC our model takes the feature maps from early stages \textbf{prematurely} and compresses them into a linear representation via FlexPooling, where no convolutional processing is involved, only completely downsampling the feature maps into singlepixel representation. We repeat this procedure after each available convolutional blocks and then take the weighted sum of these which give us our final loss which we use to train our model. For our case we designate the weights of the loss as $\lambda1$, $\lambda2$ and $\lambda3$ whose values after cross validation we choose as 0.1, 0.2 and 0.7 respectively. Learning rate use is 0.1 and we use one cycle learning rate.

\subsection{Training details }

For the case of FlexPooling method we train our models for 100 epochs with batch size 1000 for
one A100 GPU. For the case of super classing we train it two steps, in the first stage we train for 10 epochs and then in the next step we train it for 100 epochs. We use the ADAM\cite{ADAM} optimizer
with an initial learning rate 0.1. We also use augmentation of random flipping, scaling to improve the generalization ability of our model.
\subsection{Main Results}

 As shown in Table 1, we test  our model on several low scale and medium scale bench mark datasets. Where Backbone use in our case is ResNet20 we may use any other backbone. The Avg pool shows the accuracy we obtain with  standard average pooling while FlexPool indicates the accuracy obtained after the FlexPooling. Further, results with regularization  as well as with addition of drop out are also given. We can see in the Table 1 that we get an overall increase of 1.5 to 2\% across all range of datasets when we apply FlexPooling method in single classifier settings. Further we also give the results for the case when we apply FlexPooling with multiscaling i.e we take varied scale outputs after each convolutional blocks and use FlexPooling individually after each block and then average these multiscaled losses to get the final weighted loss. The results for this approach give us clear  increase of 1 - 3 \% increase across the range of all datasets.
% Further we apply with super classing approach for small scale datasets and get an increase in the range of 12\% to  20 \%.

\subsection{Ablation Study}

To further analyze the ability of our proposed  method, we conduct
extensive ablation studies on the several datasets to explore the effects of our components. We use the official
training and validation split and accumulate the evaluation
results over the whole training set. The ablation study results are shown in Table 1, Table 2 and Table 3. Our baseline model use average pooling, whereas we introduce the FlexPool and FlexPool with SAC, replacing standard average pooling to see the effects. Below we discuss the different ablation studies we perform during the course of our experiments to validate our method. 

\subsubsection{Effect of FlexPooling}
Initially we apply the FlexPooling after the final convolutional block before feeding the features maps to the final projection layer. Results in Table 1 confirm that  FlexPooling improves the performance of our model for all datsets discussed. We get the highest increase for the case of ImageNet benchmark which give us almost 0.5\% increase without any regularization and dropout. We then apply regularization for improving the generalization and we see further 1\% increase which give us 48.11 accuracy compared to the average pooling which give us 46.5 accuracy. We see that adding 25 \% dropout increases the accuracy to 48.55.  Similarly for the case of CFAR100 we get increase of 0.3 \% without regularization to 1.8\% increase with regularization and dropout. We see similar trends for the case of other datasets given in Table 1. \\For these settings we get the best results for the case of Tiny ImageNet which improves the result by almost ~2 \%. The FlexPool with regularization and using drop out of 0.25 outperforms in accuracy as it is able to learn jointly with entrire network, thus enhance its ability to exract  more meaningful feature map representations that help with model's generalizability and discriminability.

\subsubsection{Effect of FlexPooling using SAC}
We further extend our idea of FlexPooling such that instead of applying it at the end of last convolutional block before the linear projection layer, we take the outputs at different depths of model and apply the FlexPooling at these depths and then we use Cross Entropy loss (CEL) to contribute in the total loss which is comprised of average of weighted sum of all these losses.\\ Table 2 confirms the improved accuracy obtain by multiscale FlexPooling while using simple auxiliary classifiers(SAC). Here network takes feature maps from early stages and compress them into linear representation via FlexPooling, where no convolutional processing is involved,yet we downsample the feature maps into single pixel representation. also sustain the increase but it give us overall increase of 0.75 \% to  2.3\%  increase over the range of all the datasets. 

Results in Table 2 confirm that adding FlexPooling 
stages helps improve the performance of our model in different ranges. We get the best increase for the case of Tiny ImageNet dataset which give us almost 2.3\% increase without any regularization and dropout. We then apply regularization for improving the generalization and we see further 1.10\% increase which give us 48.64 accuracy compared to the average pooling which give us 46.5 accuracy. We see that adding 0.25 \% dropout and regularization weights give us nearly ~3.5\% improvement. Similarly for the case of CFAR100 we get increase of 0.75 \% without regularization to 1.1\% increase with regularization and dropout. We see similar trends for the case of other datasets given in Table 1. For these settings we get the best results for the case of Tiny ImageNet which improves the result by almost ~3.5 \%. The consistent improvement across various datasets show the stability of the method.

\section{\uppercase{Conclusions}}

% Most convolutional neural networks (CNNs) used in computer vision have
% a fundamental structure consisting of many feature extraction processing layers. The input signal is downscaled into a lower resolution in each successive layer. This downsampling procedure is known as pooling and is an essential CNN operation. It raises the receptive field, size, decreases the number of trainable parameters, strengthens the model’s resistance to 
% transformational variances, and speeds up computation.

In this paper, we present FlexPooling approach with and without simple auxiliary classifier(SAC). FlexPooling with SAC is a straightforward but efficient adaptive pooling technique that learns weighted average pooling over activations together with the rest of the network, thus generalizing the idea of average pooling with consistent improved performance.
In our approach, we make sure that each successive layer repeats the most salient information from the prior activations because pooling is a lossy operation but essential in separating high-level information from low-level information. This improves the network's discriminability. Secondly for FlexPooling with SAC our suggested network takes feature maps from early stages and compress them into linear representation via FlexPooling, where no convolutional processing is involved, yet we downsample the feature maps into single pixel representation. The loss values obtained from early stages of the CNN aid the network in learning more
abstract concepts by boosting the gradient signal throughout the entire network.
Our approach learns jointly with the entire network end to end, enhancing its ability to adapt and extract more meaningful feature, map representations that help with the model's discriminability and generalizability
We validate this claim by extensive experiments in single-classifiers as well as multi-classifier settings.  We obtain the stable, increasing accuracy trend in both settings from the average pool to flex pool. We show  further improvement in  accuracy when each of above mentioned settings are tested with three different ablation studies, including  FlexPool, FlexPool with regularization and FlexPool with regularization and dropout. Overall, FlexPool with(SAC) settings attain higher accuracies on average compared to a single classifier thanks to improved gradient signal throughout the CNN. Finally, we demonstrate that our technique consistently outperforms baseline networks in image classification across a variety of popular datasets, resulting in accuracy gains of 1-3\%.

% \cleardoublepage

% \bibliographystyle{apalike}
% {\small
% \bibliography{ref}
% }
\bibliography{ref.bib}
\bibliographystyle{apalike}
\end{document}